\documentclass[sigconf]{acmart}

\usepackage{booktabs} 
\usepackage{graphicx}
\usepackage{amsmath}
\usepackage{balance}  
\usepackage{array}
\usepackage{subfiles}
\usepackage{url}
\usepackage{array}
\newcolumntype{L}{>{\centering\arraybackslash}m{3cm}}

\usepackage{float}

\usepackage{amsmath} 
\numberwithin{equation}{subsection} 
\makeatletter
\@addtoreset{equation}{section}
\makeatother

\usepackage{multirow}

\makeindex
\begin{document}

\copyrightyear{2019}
\acmYear{2019}
\setcopyright{acmcopyright}
\acmConference[CoDS-COMAD '19]{6th ACM IKDD CoDS and 24th
COMAD}{January 3--5, 2019}{Kolkata, India}
\acmPrice{15.00}
\acmDOI{10.1145/3297001.3297029}
\acmISBN{978-1-4503-6207-8/19/01}

\title{Fast Online 'Next Best Offers' using Deep Learning}
\author{Rekha Singhal}
\affiliation{%
  \institution{TCS Research}
  \city{Mumbai}
}
\email{rekha.singhal@tcs.com}

\author{Gautam Shroff}
\affiliation{%
  \institution{TCS Research}
  \city{Delhi}
}
\email{	gautam.shroff@tcs.com}

\author{Mukund Kumar}
\affiliation{%
  \institution{TCS Research}
  \city{Mumbai}
}
\email{mkmukund10@gmail.com}

\author{Sharod Roy Choudhury}
\affiliation{%
  \institution{TCS Research}
  \city{Mumbai}
}
\email{sharod.rchoudhury@tcs.com}

\author{Sanket Kadarkar}
\affiliation{%
  \institution{TCS Research}
  \city{Mumbai}
}
\email{sskadarkar@gmail.com}

\author{Rupinder Virk}
\affiliation{%
  \institution{TCS Research}
  \city{Mumbai}
}
\email{rupinder.virk@tcs.com}

\author{Siddharth Verma}
\affiliation{%
  \institution{TCS Research}
  \city{Delhi}
}
\email{verma.sid@tcs.com}

\author{Vartika	Tewari}
\affiliation{%
  \institution{TCS Research}
  \city{Delhi}
}
\email{vartika.tewari@tcs.com}

\begin{abstract}
In this paper we present iPrescribe, a scalable low-latency architecture
for recommending 'next-best-offers' in an online setting. The paper presents the design of iPrescribe and compares its performance for implementations using different real-time streaming technology stacks. iPrescribe uses ensemble of deep learning and machine learning algorithms for prediction. We describe the scalable real-time streaming technology stack and optimised machine-learning implementations to achieve a 90th percentile recommendation latency of 38 milliseconds. 
Optimizations include a novel mechanism to deploy recurrent Long Short Term Memory (LSTM) deep learning networks efficiently. 
\end{abstract}

%
%

\maketitle

\section{Introduction}
Data analytics has evolved from descriptive, diagnostic and predictive to prescriptive analytics for effective business operations. Prescriptive analytics refers to 'what shall I do' to engage customers in Business to Consumer (B2C) systems using recommendations and/or  campaigns. Next Best Offer (NBO) is an extension of recommendation system keeping business objectives in focus. NBO refers to the 'right' offer, given to a customer at 'right' time in online setting while maximizing business objectives. iPrescribe is a high performance implementation of NBO which co-locates with B2C system.

B2C system processes millions of transactions per second with low latency and thus requires high scale. Any recommendation system interfacing with B2C system in real time necessitates to cope up with its performance and workload. Therefore, iPrescribe, needs to support high throughput and
very low (few milliseconds) latency for making recommendations model inference with high accuracy.

iPrescribe uses analytical model to predict a customer's repeat probability, which needs to be built using  machine and/or deep learning techniques on transaction data, social feeds and data from other business channels. This prediction model is used to assign offers to users while optimizing business objective functions ~\cite{iprescribe}. 

Machine learning models work best when used with the right features. Deep learning models gives higher accuracy, however, model inference 
using RNNs (such as LSTMs) requires each user's history for each prediction, leading to higher model inference time~\cite{DeeplearnShop}. Combining multiple models' predictions yields better accuracy~\cite{ensemble} than traditional collaborative filtering and segmentation techniques~\cite{collaborativefiltering2}. 

 iPrescribe uses ensemble of machine learning (gradient boosting, XGBoost~\cite{xgboost}), and deep learning (Long Short Term Memory (LSTM)~\cite{lstm}) with optimisation to avoid having to store and process user-histories for every prediction. We have achieved reasonable accuracy for the domain
 in question, as measured by  Area Under Curve (AUC, where True Positive Rates are plotted against the False Positive Rates) is 0.67 and F-Score (Harmonic mean of precision and recall) is 0.3843 for our ensemble model. The model is configurable and can be build for any business domain since iPrescribe uses meta-models, which define  business transaction data sets and functions for creating business specific features. The details on meta-model is out of scope of this paper and here, we focus only on the design, implementation and optimizations of iPrescribe system.

We have employed optimizations for XGBoost and LSTM model inference on transaction data, to reduce inference time. 
Further, performance gains in reducing recommendation latency are obtained by tuning open source stream processing architecture stack used to build iPrescribe. The use of open source modular technology stack brings agility to the iPrescribe architecture in a way that it can be built with different technology choices for each layer in the stack as discussed later in the Section~\ref{section:arch}, comparable to systems such as \cite{rs} that also use open source big data technology. The proposed architecture allows iPrescribe to be autonomous so that it can detect degradation in the system performance and scale out.
\subsection{Contributions}
Our key contributions are as follows:
\begin{itemize}
\item
Real-time high performance streaming architecture for `next-best-offer' recommendations, achieving low recommendation latency (90th percentile 38 milliseconds) and scalable high throughput using open source big data technology.
\item
Optimizations in XGBoost and LSTM algorithms implementations to reduce model inference time.
\end{itemize}
The rest of this paper is organized as follows: Section 2 presents the related work, Section 3 presents the requirements for designing iPrescribe architecture. Section 4 describes components of iPrescribe architecture using open source technology. Section 5 presents performance optimizations in iPrescribe architecture which includes both model specific and technology stack specific optimizations for design exploration. Model specific optimization subsection also discusses the accuracy of the ensemble model. Section 6 presents the performance evaluation of iPrescribe architecture on two publicly available e-commerce data sets. Finally the paper concludes in Section 7.
\section{Related Work}
Lot of work done has been done in the area of personalized recommendation~\cite{news:scalable} and campaign management~\cite{intelli}. Recommendation engines are in existence since a decade.~\cite{rs} lists the open source recommender systems.
Their focus has been on the model accuracy~\cite{recbench}. Traditionally, collaborative filtering has been used to build customer repeat probability prediction model~\cite{news:scalable, pixie, phoenix:perf}. However, predictive accuracy is substantially improved when blending multiple predictors~\cite{ensemble}, which may require continually updated features for better accuracy.

In open source recommendation system, ~\cite{Seldon} based on big data technology is quite closer to iPrescribe architecture and has shown around 99th percentile recommendation latency  as 200 ms, whereas iPrescribe shows 4 times better latency in Fig~\ref{fig:recommendthr}.
~\cite{trillionscale:features} performs multiple passes over data to build model using Hadoop technology but, iPrescribe require single pass on data to build feature dictionary which is then used to build concatenated one-hot vectors and model in lesser time.  
~\cite{youtube:model} uses bigtables to build recommendation model offline therefore features are not updated in real time unlike iPrescribe.~\cite{Sheth2013} has used caching to improve performance for recommendation system in terms of reducing latency and increasing throughput similar to iPrescribe in-memory feature store for similar performance gains.~\cite{tushar:deep} is closest to our work. They also employ ensemble of  machine and deep learning algorithms to build recommender systems. They have published around 14 ms as recommendation latency, however architecture details are not shared. Moreover, their system is not claimed to be configurable for different data sets.~\cite{aerospike} had proposed a similar requirement architecture for real time operational DBMS.

\section{Design Requirements}\label{section:req}
\begin{figure}[ht]
\centering
\includegraphics[scale=0.4]{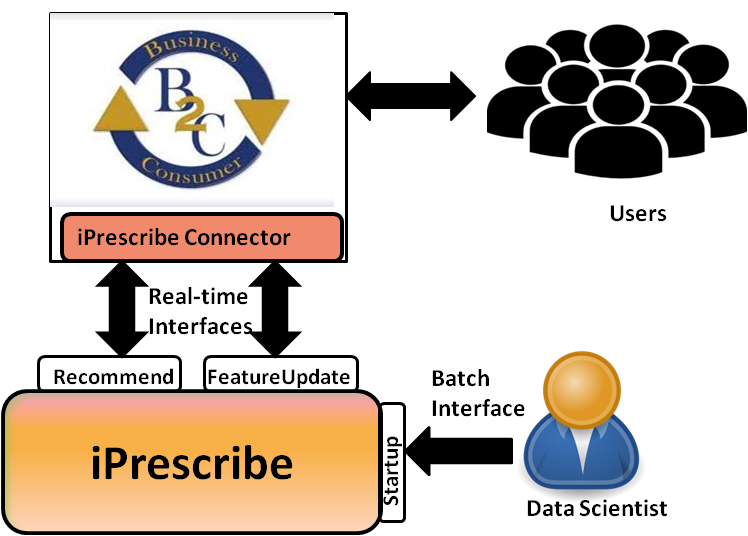}
\caption{iPrescribe with B2C System}\label{fig:bigpic}
\vspace{-4mm}
\end{figure}

For real time deployment of iPrescribe, on-line users' actions on B2C system are processed and applied on the already built model in real time for deciding the right offer. Therefore, iPrescribe, as shown in Fig~\ref{fig:bigpic} requires two types of interfaces: Batch and Real-time.
In this paper, we consider ensemble of LSTM and XGBoost algorithms in Keras and Python\footnote{Python is chosen being most popular among data scientists} respectively to build customer repeat probability prediction model. Each of these techniques requires creating features from transaction data and then building respective models. Transaction data depicts temporal behaviour of users which is captured in features and need to be updated continuously. For example, a feature indicating a "user's buying behaviour in the last two hours" will be different in morning and evening of a day. The current value of such features improves the accuracy of XGBoost algorithm. These features are kept updated by getting all users' actions from B2C system through one of the real time interfaces of iPrescribe. 
\subsection{Interfaces Performance Metrics}\label{section:interface}
\begin{figure}
\centering
\includegraphics[height=2.9in, width=3.5in]{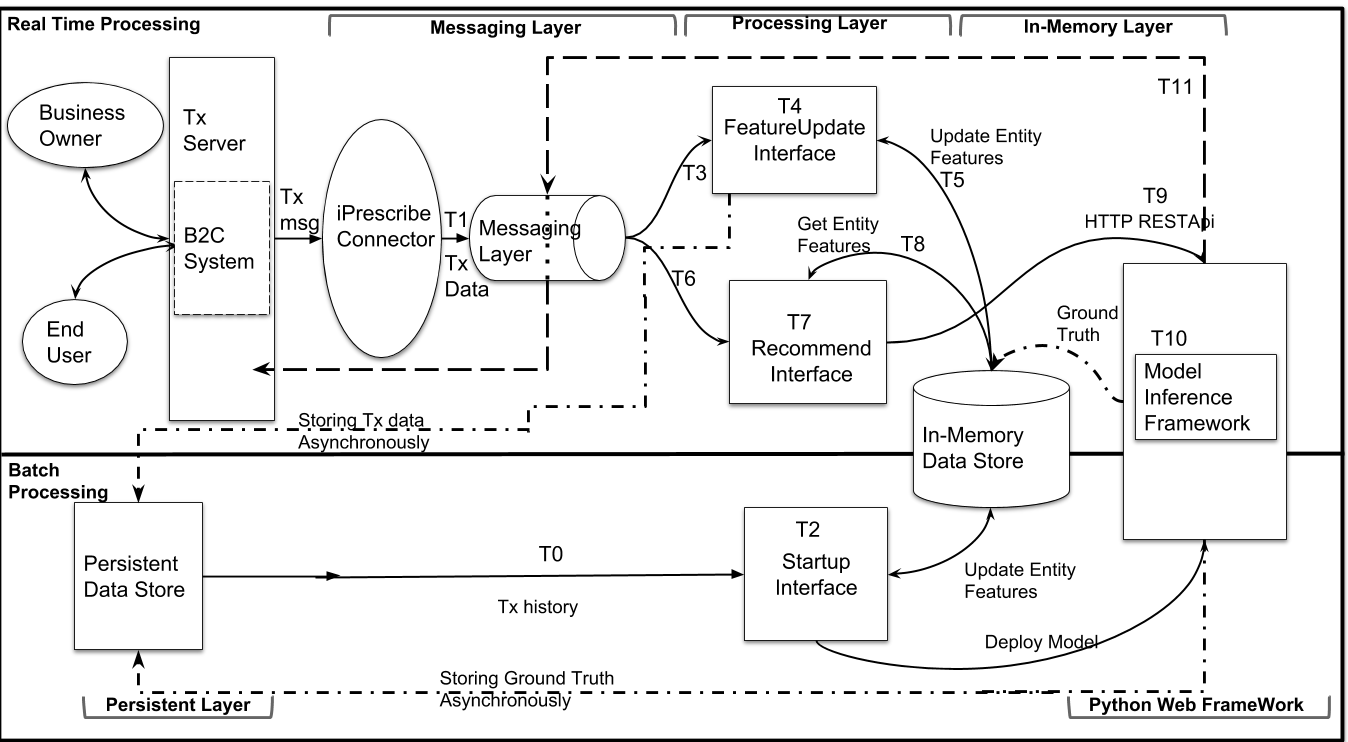}
\caption{iPrescribe Interfaces Processing}
\label{fig:interfaces}
\vspace{-4mm}
\end{figure}

iPrescribe, as shown in Fig~\ref{fig:bigpic} requires two types of interfaces,batch interface to train the XGBoost and LSTM models and real time interfaces to capture users' on-line actions and/or transactions on B2C system and return relevant offers. Thus iPrescribe has three interfaces. A batch interface, 'Startup', to build model from transaction data and deploy in memory. Two real time interfaces, 'Recommend' and 'FeatureUpdate' for getting best offer for user and updating model features respectively.
Fig~\ref{fig:interfaces} shows the detail processing involved at each layer of the architecture stack for these interfaces, with timing notations 'Tx' refer to time required by each component for processing and also time taken for data transfer from one component to another.
\subsubsection{Startup Interface}
This interface performs the complete pipeline of  building model from transaction data: feature creation, training and testing the model. The feature creation process reads transaction data file and builds concatenated feature one-hot vectors which are also stored in data store. These concatenated one-hot vectors for each target (e.g. user) are used to build the XGBoost model. Similarly, LSTM model is built using the transaction sequences for each user. The performance metric for this interface is model building time which includes time to create the concatenated feature one-hot vectors and then training the model, this is 
referred as $Execution Time$.
\begin{equation} Execution Time = T0+T2 \end{equation} 
\subsubsection{Recommend Interface} 
This real time interface is invoked through iPrescribe connector for every action message which need to trigger an offer for a user. This is a closed loop system, therefore, the performance metric of interest for this interface are \textit{recommendation latency (RL)} and \textit{throughput}. This involves retrieving the user's context from an incoming message, fetching its features for both the models, preparing the concatenated feature one-hot vectors for XGBoost and input vector/matrix for LSTM models, ensemble inference of both the models to get the best offer for the user. Finally, the assigned offer is sent back to the B2C system through the connector. 
$RL$ is time taken to send offer to the B2C system once iPrescribe receive a message.
\begin{equation} RL = T1+T6+T7+T8+T9+T10+T11,\end{equation}\label{eq:rl}
Throughput is measured as number of messages serviced per second. Ideally, throughput shall increase linearly with increase in message ingestion rate till the system is fully utilized. Therefore,
\begin{equation} Throughput= Min \left\{\frac{cores_{msg}}{T7+T8},  \frac{cores_{infr}}{T10}\right\} \end{equation}
where $cores_{msg}$ and $cores_{infr}$ are number of cores at message processing and model inference layers respectively.
\subsubsection{FeatureUpdate Interface} This real time interface is invoked through iPrescribe connector for every action on the B2C system. Its purpose is to keep the features updated for every action. This is an open system therefore, the performance metric for this interface is only \textit{throughput}. The workflow involves retrieving user details from an incoming message, fetching the user's existing features, updating the features with current context and storing it back in the feature store.  Throughput is measured as the  maximum number of messages serviced per second while maximally utilizing the underlying system.
\begin{equation}Throughput = (1/ProcessTime) * Numcores\end{equation} where $Numcores$ is number of cores in stream processing layer and 
\begin{equation} ProcessTime=T1+T3+T4+T5 \end{equation}\label{eq:pt}
\section{Architecture}\label{section:arch}
The performance of real time interfaces depend on the feature store access time, processing time and model inference time. 
The key components of the architecture are the design of feature storage for faster access and the technology stack encapsulating multiple layer architecture as discussed in this section.
\subsection{Feature Storage Structure}\label{section:featurestore}
Features are created for both XGBoost and LSTM model by processing transaction history. For the XGBoost model, to capture user's persona and temporal behaviour, we define two categories of features: non-temporal and temporal, respectively for each user, which is similar to JSON~\cite{json} data type. We reduce feature access time by sharding and creating indexes on 'user id', so data for a user can be accessed in O(1) rather of sequential scan. Each user's features are calculated in two passes. 

The first pass on transaction history creates features' cumulative values for each user, e.g. total count of product view in last 3 days, which is referred to as 'feature dictionary'. The second pass on  the features' cumulative values creates feature one-hot vectors for users, e.g. favorite category of user, and then concatenates them, which is referred to as 'concatenated feature one-hot vectors'. To reduce the processing time at 'Recommend' interface, both feature dictionary and the concatenated feature one-hot vectors are kept in in-memory store.
\subsection{Technology Stack Choices}
We have used python to build models used in iPrescribe, so a naive approach is to build model offline and deploy it using python web based frameworks (PWF)~\cite{pwf} such as Flask~\cite{flask} and Tornado~\cite{tornado}. iPrescribe connector could capture on-line activities of B2C system  and send it to PWF to get model inference. Python being an interpreter, this architecture will have challenges. The challenges include large disk access time, scalability, impact on B2C system performance.
Thus, iPrescribe is implemented as five layer architecture stack outside B2C system.
\subsubsection{Message and Persistant Store Layers}
The open source technology, Kafka~\cite{kafka}, and Hadoop Distributed File System (HDFS)~\cite{hdfs}, are considered for  horizontally scalable message layer and persistent store respectively. All the actions and transactions of users on B2C system are captured as real time messages through real time interfaces. These are stored asynchronously in persistent store for future model rebuilding. Received messages are also co-related with actual conversions for a given offer and are asynchronously stored as ground truth both in persistent and in-memory store for model rebuilding and updating feature store in real time respectively.
\subsubsection{In-memory Store Layer}
The feature store schema depends on the data sets, therefore, technology for in-memory layer must support dynamic schema creation and JSON data types. 'Recommend' and 'FeatureUpdate' interfaces will only access the feature store concurrently for reading and  updating only respectively, therefore, iPrescribe data store need not have strong transaction consistency. 'Recommend' interface may read feature values without reflecting updates of few recent actions which may not impact model inference accuracy. We explored Mongo DB~\cite{mongo} and Ignite~\cite{ignite} for in-memory store and their impact on performance is discussed later in Section~\ref{section:perfopt}.
\subsubsection{Stream Processing Layer}
For scalable parallel data processing, Spark~\cite{spark} and ignite~\cite{ignite} are explored. Spark supports python, as PySpark~\cite{pyspark}, but has no memory store and Ignite does not support python but has in-memory store. Spark being Java based, it has additional python workers which lead to double serialization overheads. Moreover, Spark is a micro batch stream processing engine, therefore $RL$ is bounded by the batch window size. Ignite supports per message processing and is a single technology for both stream processing and in-memory store; this reduces the message processing time to few milliseconds only. This is discussed in detail later in Section~\ref{section:experiment}. 
\subsubsection{Python Web Framework Layer (PWF)}
Python web framework is used only for model inference. Real time messages are processed in parallel by stream processing layer and sent to PWF for model inference using HTTP RestAPI calls. Each python process executes independent of any other process, therefore, PWF layer can be scaled out with more resources, upon increase in workload to ensure constant model inference time. 
\section{Performance Optimizations}\label{section:perfopt}
This section discusses tuning of iPrescribe architecture. This includes XGBoost and LSTM model specific optimizations and the parameter tuning of various technology stacks for design exploration to achieve high scale iPrescribe with low recommendation latency.
\subsection{Model Specific Optimizations}\label{section:modelopt}
We have employed ensemble of XGBoost and LSTM algorithms to build the prediction model. Model inference time can be reduced by batching users' concatenated feature one-hot vectors. It implies that messages coming to 'Recommend Interface' within a few milliseconds can be processed in parallel and send to PWF together for model inference.
\subsubsection{Model Building}\label{section:model}
A transaction history captures static information about entities and dynamic information about actions or transactions on entities. 
We have used PAKDD Recobell challenge~\cite{pakdd} and Kaggle Instacart challenge~\cite{instacart} datasets, details given in Table~\ref{table:datasets}, to build the model. 
\begin{table}
\centering
\caption{Statistics of Data Sets }
\begin{tabular}{|p{2.5cm}|p{2.5cm}|p{2.5cm}|}\hline
\textbf{Statistic} & \textbf{PAKDD Recobell} & \textbf{KAGGLE Instacart} \\ \hline
\# event samples & \multicolumn{1}{m{2.5cm}|}{4,80,26,835} & \multicolumn{1}{m{2.5cm}|}{34,21,083}\\ \hline
\# users & \multicolumn{1}{m{2.5cm}|}{21,18,678 } & \multicolumn{1}{m{2.5cm}|}{2,06,209} \\ \hline
\# products & \multicolumn{1}{m{2.5cm}|}{4,22,880} & \multicolumn{1}{m{2.5cm}|}{ 49,688}\\ \hline
Dataset duration & \multicolumn{1}{m{2.5cm}|}{ 1 Aug,'16 - 1 Oct,'16} & \multicolumn{1}{m{2.5cm}|}{ 1 Year}\\ \hline
Imbalance in target &  \multicolumn{1}{m{2.5cm}|}{10\% positive class} &  \multicolumn{1}{m{2.5cm}|}{4\% positive class} \\ \hline
\end{tabular}\label{table:datasets}
\vspace{-7mm}
\end{table} 
 
Our machine learning  model uses Gradient Boosting(XGBoost), where grid search was used to select the optimal parameter values for following XGBoost model parameters - colsample\_bylevel, colsample\_bytree, learning\_rate, max\_depth, min\_child\_weight,  n\_estimators and subsample.
Deep learning algorithm can encapsulate many hidden features which cannot be captured using programmed feature engineering. We have used LSTM deep neural network. The model structure has 150 node and 20 node LSTM layer for PAKDD Recobell and Instacart respectively, followed by a dense layer with 2 nodes and a softmax activation function. We have also used l2 regularizer and rms prop as an optimizer with a learning rate of 0.001. 
We apply a weighted ensemble of predictions from gradient boosting algorithm as well as LSTM model to cover the spectrum of features which together can improve the accuracy. We calculate the weights given to predictions of both the algorithms to optimize the Area Under Curve (AUC). Threshold function is applied on the probabilities obtained after the ensemble to optimize the F-Score on the final predictions. We obtained Area Under Curve (AUC) score of 0.67 on PAKDD dataset and F-Score of 0.3843 on Instacart dataset.


\subsubsection{LSTM Optimizations}\label{section:lstmopt}
LSTM, being a sequence based model,  the model inference requires passing the whole sequence of transaction history to the network architecture. This leads to large model inference time which may increase over a period of time with increase in number of sequences. 
Naive approach of LSTM model inference technique takes 36 hours to train 22 million records and take 831 ms for a user with history of 10,000 samples. This is due to looping back of last hidden states and cell states for new sequence vector. The looping back of network can be unfolded as multiple sequence of  the LSTM units~\cite{lstm}, each feeding to the next in sequence. 
According to the equations of LSTM~\cite{lstm} only $h_{t}$ and $c_{t}$ are passed to the next time step. At any point of time the values of $h_t$ and $c_t$ together represents the LSTM network state trained with the historical data till 't-1'. Therefore, LSTM model inference for a message at time 't' can be done in constant time on LSTM network loaded with value of $h_{(t-1)}$ and $c_{(t-1)}$.~\cite{LSTMsame} has used $h_t$ for LSTM model inference, however they have not exploited it for performance gains in real time model inference.

iPrescribe feature store stores $h_t$ and $c_t$ for each user as well during model training in 'Startup' interface. These values are updated for each message in 'FeatureUpdate' interface.
Moreover, Keras library predict function incurs 67 ms and 25 ms for model inference using small size with more matrix multiplications and large size with less matrix multiplications respectively, to predict for one user; most of the time is taken up by core tensorflow back-end built-in methods TF\_ExtendGraph and TF\_Run, and other internal calls of tensorflow. iPrescribe has its own implementation of 'predict' function in Java using JBLAS 1.2.4~\cite{jblas}, LAPACK and ATLAS for optimized matrix multiplication. The mapping of categorical columns to unique integers is done using hashing instead of data store to reduce inference time further.
\subsection{Design Exploration with Technology Stack Optimizations}\label{section:techopt}
iPrescribe architecture as discussed in Section~\ref{section:arch}, has been implemented for various technology choices given in Table~\ref{table:techchoice} to explore high scale and low latency technology stack architecture. 
\begin{table}[H]
\centering
\caption{Technology Choices for iPrescribe} 
\begin{tabular}{|l|c|} \hline
\bf{Architecture Layer}&\bf{Technology}\\ \hline
 Messaging Layer & Kafka\\ \hline
 Stream Processing Layer & Ignite \\ \hline
 In-memory Store & Ignite\\ \hline
 Python Web Framework & Tornado\\ \hline
 Persistent Store Layer & HDFS\\ \hline
 \end{tabular} \label{table:techchoice}
 \vspace{-4mm}
 \end{table}  
Kafka, the messaging layer, is partitioned with multiple topics to support higher ingestion rate, however, performance gains with increase in number of partitions are limited by disk access overheads in Kafka for messages persistence.
The performance optimizations in rest of the technology stack is discussed below.
\begin{figure}
\centering
\includegraphics[height=0.8in, width=3.3in]{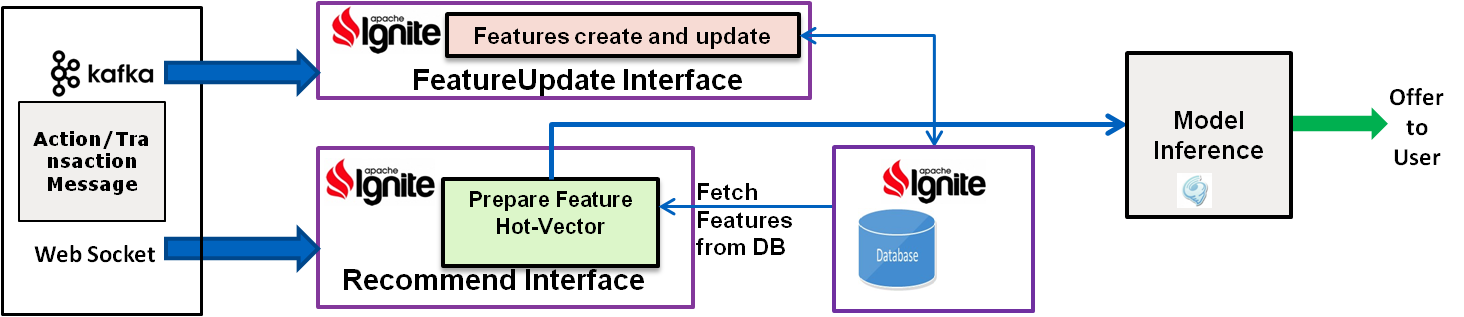}
\caption{IP Architecture}\label{fig:ignite_arch}
\vspace{-4mm}
\end{figure}
The model is hosted as web service on Tornado server, with facility of multiple processes, which listens to a particular port. However, using same port to serve larger number of requests will limit the throughput. Therefore, multiple Tornado processes are started independently, each one listening to different port. 

Multiple process deployment of PWF (Tornado) leads to sub linear speed up with vertical scaling of a machine and hence increases the model inference latency unlike single server which incurs only 12 ms, i.e. the system is not scalable. This is because the XGBoost library spawns multiple threads and these threads waits on python GIL. We could limit number of threads to one for XGBoost by setting environment variable OMP\_NUM\_THREADS=1. 
Further, to avoid context switching on cores across different Torando processes, each Tornado process is attached to a core using environment variable 'Task set'.
In this architecture after optimization, average \textbf{$RL$ is 78 ms}, where time taken for message processing in Spark is 50 ms including 28 ms for time spent in accessing MongoDB, which is quite high. So we have used Ignite, and call our architecture to be Ignite+PWF (IP) Architecture.
Ignite is both per message stream processing and in-memory store technology. Ignite Cache, key value store, is used to store features. The architecture is shown in Fig~\ref{fig:ignite_arch}. User identifier is used as the key, to store and partition data across nodes in the Ignite cluster. Ignite StreamVisitor is being used for processing each key-value tuple from incoming data streams. In Ignite cluster, StreamVisitor collocates processing locally on the node where the data is cached to avoid data shuffling.
This reduces message processing time to 14 ms. However, Ignite-Kafka connector introduces overheads of 45 ms. Therefore, web sockets are used to send messages from iPrescribe connector to Ignite client which reduces the communication delay to 2 ms with the tradeoff of Kafka's reliability and availability. In this architecture after optimization, average \textbf{$RL$ is 29.52 ms}.
To support high ingestion rate, multiple Ignite client instances are launched with every instance listening to separate web socket. \\
\section{Experimental Evaluation}\label{section:experiment}
In this section, we will discuss performance evaluation of iPrescribe optimized architecture. 
\subsection{Deployment System Details}
We have deployed the technology stack of iPrescribe on six node cluster, each node with the following configuration: Intel CPU dual core with 56 cores and 256 GB RAM with 1 GB NIC. 
Choice of technology while design exploration of iPrescribe architecture and their deployment details with tuning is given in Table~\ref{table:techchoice} and Table~\ref{table:deploy-tune} respectively. Kafka and MongoDB are low on resource utilization in this case, therefore they are deployed on shared nodes. 
\begin{table}[]
\centering
\caption{Technology Deployment and Tuning} 
\begin{tabular}{|l|l|L|} \hline
\bf{Technology} & \bf{Deployment} & \bf{Tuning} \\ \hline
Kafka 2.10 & 3 nodes cluster & \multicolumn{1}{m{2.6cm}|}{10 partitions at each node}\\ \hline
Ignite 1.9 & 2 nodes cluster & \multicolumn{1}{m{3.2cm}|}{Cache Memory= Off-Heaped, Indexing on UserId, Java Heap Size of Server=4GB, Cache-Partition= UserId} \\ \hline
HDFS 2.6 & 1 node & \multicolumn{1}{m{2.6cm}|}{Default} \\ \hline
Tornado 4.5 & 2 shared nodes & \multicolumn{1}{m{2.6cm}|}{Default} \\ \hline
 \end{tabular} \label{table:deploy-tune}
 \vspace{-5mm}
 \end{table}   
\subsection{Benchmark Workload}
iPrescribe architecture performance is evaluated on PAKDD~\cite{pakdd} and Instacart~\cite{instacart} data sets. PAKDD has 22 million transaction records including 0.3 million impression records. The built model predicts whether a customer will click the given advertisement. Instacart has 2,06,209 users with 50,000 products, where the built model predicts whether a particular customer will buy a particular product. The model can be used for all products, to predict products in a customer's basket for next order. The model is built on the initial data set in Startup Interface discussed in Section~\ref{section:interface}.  We have extrapolated the impression records of PAKDD, to generate large number of impression records. These impression records are played as stream and fed to real time interfaces of iPrescribe. These records are sorted on clock time, so it simulates the behaviour of user clicks on transaction system. Similarly, test data of Instacart consisting of user records is simulated as stream to benchmark real time interfaces.  We control the ingestion rate of the records and measure system throughput and utilization. For example, for PAKDD data sets, Recommend interface is ingested with stream of impression records, and FeatureUpdate interface is ingested with stream of records having mix of all order, view and impression records. We have benchmark iPrescribe for only 100\% workload on Recommend interface, 100\% workload on FeatureUpdate and controlled ingestion rate on both the interfaces in ratio of 80\% and 20\% respectively on FeatureUpdate and Recommend interfaces.
\subsection{Performance Results}
Startup Interface reads CDF file, prepares users' concatenated feature one-hot vectors  in parallel by processing transaction records using meta-model and builds ensemble of XGBoost and LSTM model in python framework. The execution time for this interface is 183 minutes and 269 minutes for PAKDD and Instacart data sets respectively. The details are given in Table~\ref{table:exetime} for feature creations and training time for each of the model.
XGBoost model for Instacart is trained for 10x less number of users than that of PAKDD model, therefore XGBoost startup interface execution time is lesser in Instacart. LSTM model requires history of each sample during training and LSTM model for Instacart is trained for all user-product pairs which is 30x more than that in PAKDD, which lead to higher LSTM model training time.
\begin{table}
\centering
\caption{Startup Interface Execution Time (in minutes)} 
\begin{tabular}{|l||c|c|c|c|} \hline
\bf{Data Set} & \multicolumn{2}{|c|}{\bf{XGBoost}} & \multicolumn{2}{|c|}{\bf{LSTM}}  \\ \hline
 & \bf{Features} & \bf{Training} & \bf{$h_t$,$c_t$} & \bf{Training}  \\ \hline
\bf{PAKDD} & 90 & 6 & 20 & 67  \\ \hline
\bf{Instacart} & 6 & 21 & 12 & 230 \\ \hline
 \end{tabular} \label{table:exetime}
 \vspace{-3mm}
 \end{table} 

\begin{table}
\centering\
\caption{PAKDD: Average $RL$ Timings as shown in Fig~\ref{fig:interfaces}} 
\begin{tabular}{|l|c|c|c|c|c|c|c|c|c|} \hline
\bf{Stack} & \bf{T4} & \bf{T5} & \bf{T6} & \bf{T7} & \bf{T8} & \bf{T9} & \bf{T10} & \bf{T11} \\ \hline
\bf{IP}  & \bf{7.3} & \bf{0.5} & \bf{0.5} & \bf{14} & \bf{0.5} & \bf{0.5} & \bf{10.52} & \bf{3} \\ \hline
 \end{tabular} \label{table:pakddlatency}
 \end{table}
 \begin{figure}
\centering
\includegraphics[height=1.5in, width=2.5in]{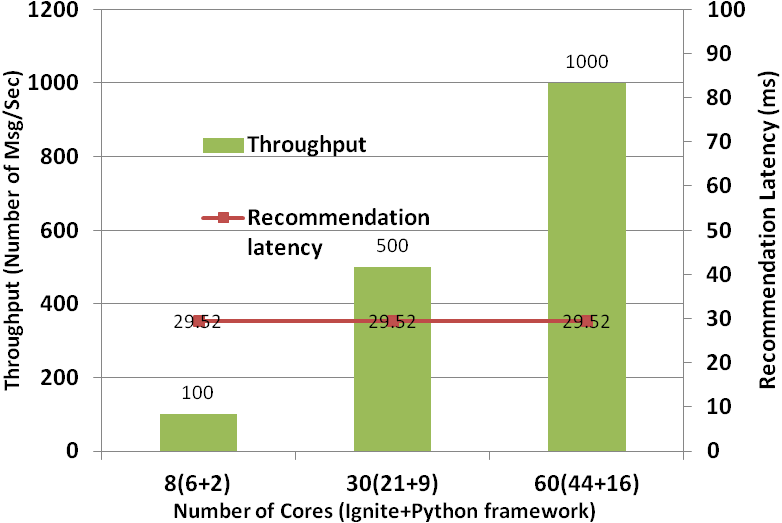}
\caption{iPrescribe Performance for Recommend Interface }\label{fig:recommendthr}
\vspace{-5.7mm}
\end{figure}
Recommend interface has two performance metrics - recommendation latency ($RL$) and throughput as given in Section~\ref{section:interface}. 
$RL$ is measured on single node starting with 100 msg/sec as ingestion rate and gradually increasing to 1000 msg/sec. In our experimental setup, $RL$ time components, $T1$=$T3$=$T6$=1 ms.
For Instacart data sets, model inference time per user-product inference is similar,  however, model inference need to be done for all user-product pairs to predict user's basket. Therefore, for an average of 40 products per user in basket, time to fetch concatenated feature one-hot vectors for XGBoost model is 81 ms and for LSTM model inference is 161 ms in stream processing layer. 

\begin{figure}
\centering
\includegraphics[height=1.7in, width=2.9in]{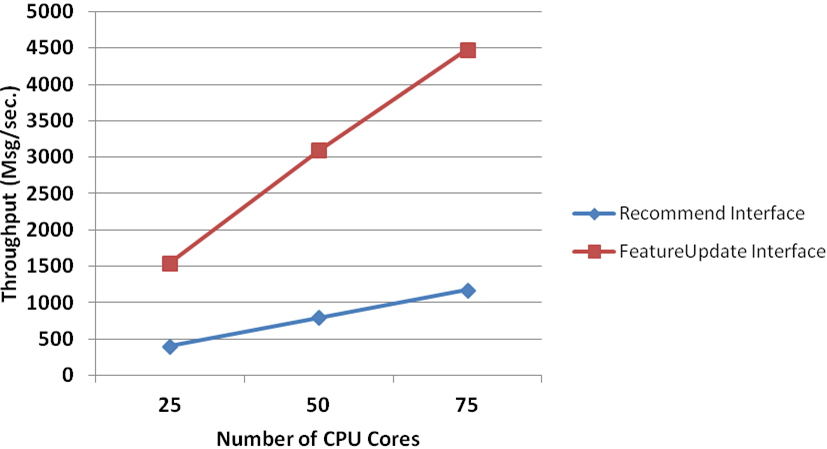}
\caption{iPrescribe Throughput for Mix workload}\label{fig:mix}
\vspace{-5.7mm}
\end{figure}

This implies, $T7$=81+161=242 ms and XGBoost model inference time in PWF per user is $T10$=480 ms. $T7$ does not increase linearly opposed to $T10$. The feature store is indexed on user id, so single fetch from Ignite Cache gets all concatenated feature one-hot vectors for a user on all products. Using equation~\ref{eq:rl}, the average recommendation latency per user for PAKDD and Kaggle Instacart challenges are 29.52 ms and 842 ms respectively.
Fig~\ref{fig:recommendthr} shows throughput and average recommendation latency in iPrescribe architecture with increase in data ingestion rate. We see linear increase in throughput with increase in workload. However, the recommendation latency remains the same as we increase the ingestion rate.
However, our experiments have shown that Ignite technology supports linear increase in throughput till the CPU utilization of the cluster is 80\%, as we go on increasing the message ingestion rate. There is a sharp increase in processing time of each message in milliseconds after the CPU utilization hits 80\%. Therefore, for high scale iPrescribe, the Ignite layer shall scale out on 80\% utilization.
FeatureUpdate interface processing time, using equation~\ref{eq:pt} and Table~\ref{table:pakddlatency}, is 7.3 ms which includes 6 ms for updating $h_t$ and $c_t$ for LSTM model and 1.3 ms for updating feature store. FeatureUpdate processing is done in Ignite in parallel across all the available cores, therefore, FeatureUpdate throughput linearly increases with number of cores and data ingestion rate, as shown in Fig~\ref{fig:mix}. It also shows that throughput of Recommend interface does not degrade in presence of processing on FeatureUpdate interface.
\section{Conclusions and Future Work}
We have presented design and different technology stacks for iPrescribe, a scalable and low recommendation latency system for next best offers in an online settings for B2C scenarios. We have shown the performance of iPrescribe on two publicly available data sets. The prediction model has been built as ensemble of XGBoost and deep learning LSTM network. These models accuracy has been shown to be AUC=0.67 and F-score=0.3843 for these two publicly available e-commerce data sets. We have discussed  model specific optimizations and tuning of various technology stacks to achieve low recommendation latency. LSTM network deployment is optimized by storing $h_{t}$ and $c_{t}$ for inference, which are updated with each transaction message. iPrescribe optimized architecture using technology stack of Kafka, Ignite, Tornado and HDFS is shown to support high scale throughput and 90th percentile recommendation latency as 38 ms. 

iPrescribe needs ground truths (or labelled data) to build model so it has problem of cold start. In future, we shall be augmenting iPrescribe with user behaviour model to capture their persona and Snorkel~\cite{snorkel} to generate labels for data sets to build models. 
\bibliographystyle{ACM-Reference-Format}
\bibliography{main}

\end{document}